%% file: main.tex
\documentclass{article}
\usepackage{spconf,amsmath,graphicx}

\usepackage[colorlinks=true,allcolors=black]{hyperref}
\usepackage{url}
\usepackage{textcomp} 
\usepackage{gensymb} 
\usepackage{arydshln}
\usepackage{booktabs}
\usepackage{multirow}
\usepackage[utf8]{inputenc}
\usepackage[acronym]{glossaries}
\usepackage[caption=false]{subfig}
\usepackage[table,dvipsnames]{xcolor}
\usepackage{xspace}
\usepackage[sort&compress,numbers]{natbib}
\usepackage{balance}

\usepackage{amssymb}
\usepackage{latexsym}
\usepackage{url}
\usepackage{enumitem}

\newacronymstyle{long-short-br}
{%
  \GlsUseAcrEntryDispStyle{long-short}%
}%
{%
  \GlsUseAcrStyleDefs{long-short}%  
}
\setacronymstyle{long-short-br}

\title{\Large Deep Representations for Cross-spectral Ocular Biometrics}

\makeatletter
\def\@name{ \emph{Luiz~A.~Zanlorensi\textsuperscript{1}, Diego~R.~Lucio\textsuperscript{1}}  \\[0.1ex] \emph{Alceu~S.~Britto~Jr.\textsuperscript{2}, Hugo~Proença\textsuperscript{3}, David~Menotti\textsuperscript{1}}\thanks{\scriptsize{This paper is a postprint of a paper submitted to and accepted for publication in \emph{IET Biometrics} and is subject to Institution of Engineering and Technology Copyright. The copy of record is available at the \emph{IET Digital Library}.}} \\}
\makeatother

\address{\textsuperscript{1}Department of Informatics, Federal University of Paraná, Curitiba, Brazil\\\textsuperscript{2}Postgraduate Program in Informatics, Pontifical Catholic University of Paraná, Curitiba, Brazil\\\textsuperscript{3}IT: Instituto de Telecomunicações, University of Beira Interior, Covilhã, Portugal\\[0.5ex]
\normalsize
\textsuperscript{1}\textit{\{lazjunior, drlucio, menotti\}@inf.ufpr.br} \qquad \textsuperscript{2}\textit{alceu.junior@pucpr.br}\qquad \textsuperscript{3}\textit{hugomcp@di.ubi.pt}}

\begin{document}
\ninept
\sloppy
\maketitle
\input{text/acronyms.tex}
\input{0-abstract.tex}
\input{text/introduction.tex}

\input{text/related.tex}

\input{text/methodology.tex}

\input{text/protocol.tex}

\input{text/results.tex}

\input{text/conclusion.tex}

\section*{Acknowledgment}
This work was supported by grants from the National Council for Scientific and Technological Development (CNPq)(Nos.~428333/2016-8, 313423/2017-2 and 306684/2018-2), and the Coordination for the Improvement of Higher Education Personnel (CAPES), and also gratefully acknowledge the support of NVIDIA Corporation with the donation of the Titan Xp GPU used for this research.
The fourth author work is funded by FCT/MEC through national funds and co-funded by FEDER - PT2020 partnership agreement under the projects UID/EEA/50008/2019 and POCI-01-0247-FEDER-033395.

%\footnotesize
\scriptsize
\balance
\setlength{\bibsep}{3pt}
\bibliographystyle{IEEEbib}
\bibliography{paper}

\end{document}

%% file: text/acronyms.tex
\newacronym{cnn}{CNN}{Convolutional Neural Network}
\newacronym{eer}{EER}{Equal Error Rate}
\newacronym{far}{FAR}{False Acceptance Rate}
\newacronym{frr}{FRR}{False Rejection Rate}
\newacronym{nir}{NIR}{near-infrared}
\newacronym{vis}{VIS}{visible}
\newacronym{pfb}{PFB}{pairwise filter bank}
\newacronym{svm}{SVM}{Support Vector Machine}
\newacronym{roc}{ROC}{Receiver Operating Characteristic}

\newcommand{\polyu}{PolyU\xspace}
\newcommand{\crosseyed}{Cross-Eyed\xspace}
\newcommand{\supplementary}{\url{https://web.inf.ufpr.br/vri/databases/iris-periocular-coarse-annotations/}}

%% file: 0-abstract.tex
\begin{abstract}
\textit{One of the major challenges in ocular biometrics is the cross-spectral scenario, i.e., how to match images acquired in different wavelengths (typically \gls{vis} against \gls{nir}).
This article designs and extensively evaluates cross-spectral ocular verification methods, for both the closed and open-world settings, using well known deep learning representations based on the iris and periocular regions.
Using as inputs the bounding boxes of non-normalized iris/periocular regions, we fine-tune \gls{cnn} models (based either on VGG16 or ResNet-50 architectures), originally trained for face recognition.
Based on the experiments carried out in  two publicly available cross-spectral ocular databases, we report results for intra-spectral and cross-spectral scenarios, with the best performance being observed when fusing \emph{ResNet-50} deep representations from both the periocular and iris regions.
When compared to the state-of-the-art, we observed that the proposed solution consistently reduces the \gls{eer} values by $90\%$ / $93\%$ / $96\%$ and $61\%$ / $77\%$ / $83\%$ on the cross-spectral scenario and in the PolyU Bi-spectral and Cross-eye-cross-spectral datasets.
Lastly, we  evaluate the effect that the "deepness" factor of feature representations has in recognition effectiveness, and - based on a subjective analysis of the most problematic pairwise comparisons - we point out further directions for this field of research.}
\end{abstract}

%% file: text/introduction.tex
\section{Introduction}
\label{sec:introduction}

\glsresetall

Iris recognition using \gls{nir} wavelength images acquired under controlled environments can be considered a mature technology, which proved to be effective in different scenarios~\cite{Proenca2012}.
In contrast, performing iris recognition in uncontrolled environments and at \gls{vis} wavelength is still a challenging problem ~\cite{proenca2005, proenca2010}.
Some of the latest researches consist of biometrics recognition on cross-spectral scenarios, i.e., using images of eyes from the same subject obtained at the \gls{vis} and \gls{nir} wavelengths~\cite{Hosseini2010, Sharma2014, Nalla2017, Algashaam2017}.

Recently, machine learning techniques based on deep learning have been achieving great popularity due to the results reported in the literature, which advance the state-of-the-art in various problems, such as speech recognition~\cite{Hinton2012, Zhang2017, Kim2017}, natural language processing~\cite{Glorot2011, Socher2011}, digit and character recognition~\cite{laroca2018robust, hochuli2018, laroca2019efficient} and face recognition~\cite{Omkar2015, Cao2017}.
In the field of ocular biometrics,  using deep learning representation has been advocated  both for the periocular~\cite{Luz2018, Proenca2018} and iris~\cite{Liu2016, Gangwar2016, Al-Waisy2017, Nguyen2018, Proenca2017, Nalla2017, Wang2019, Zanlorensi2018} regions,  with interesting and promising results being reported.

As stated in previous works~\cite{demarsico2016, Liu2016}, an often and open problem in ocular recognition is the matching heterogeneous images captured at different resolutions, distances and devices (cross-sensor and cross-spectral).
Regarding these problems it is difficult to design a robust handcrafted feature extractor to address the intra-class variations present in this scenarios.
In this sense, several recent works demonstrate that deep representations report better results compared to handcrafted features in iris and periocular region recognition~\cite{Liu2016, Luz2018, Proenca2018, Wang2019}.

Having in mind that deep learning frameworks are typically able to produce robust representations, in this article we apply this family of frameworks to extract and combine features from the ocular region, obtained at different wavelengths, e.g., \gls{vis} and \gls{nir}.
The strategy described in this article is composed of some methodologies extracted from the literature. For both the iris and ocular traits we use as input the bounding box delimited regions used in the state-of-the-art methods~\cite{Zanlorensi2018,Luz2018}.
Then, the features from these traits were extracted using a similar approach proposed by~\cite{Zanlorensi2018}.
In this direction, the main contribution of this article is the extensive experiments on two datasets comparing iris, periocular, and fusion results for both cross-spectral (\gls{vis} to \gls{nir}) and intra-spectral (\gls{vis} to \gls{vis}, \gls{nir} to \gls{nir}) matching, reaching a new state-of-the-art results. 
There is also the following four-fold contributions: (i)~we show that deep learning  yield robust representations on two well-known cross-spectral databases (\polyu and \crosseyed) for ocular verification using closed- and open-world protocols; (ii)~we report how two off-the-shelf networks can be fine-tuned from the face domain to the periocular and iris one; (iii)~we analyze the use of a single deep representation extraction schema, for both cross-spectral and the same spectra scenarios; and (iv) we conclude about the benefits of fusing the periocular and iris representations to improve the recognition accuracy.

The remainder of this work is organized as follows.
In Section~\ref{sec:related}, we describe some recent works that use deep learning for iris and periocular recognition.
Section~\ref{sec:methodology} provides the details of the proposed approach.
Section~\ref{sec:protocol} presents the databases, metrics and evaluation protocol used in our empirical evaluation.
The results are presented and discussed in Section~\ref{sec:results}.
Lastly, the conclusions are given in Section~\ref{sec:conclusion}.

%% file: text/related.tex
\section{Related Work}
\label{sec:related}

This section surveys the works that use deep learning frameworks for iris and periocular recognition. Also, we summarize the most relevant ocular recognition methodologies focused on the cross-spectral scenario.

One of the first works applying deep learning to iris recognition only was the \emph{DeepIris} framework,  proposed by Liu et al.~\cite{Liu2016}.
Having as a goal the recognition of heterogeneous irises using images obtained by different sensors (i.e., the cross-sensor scenario),  the authors proposed a framework that establishes the similarity between a pair of iris images using \glspl*{cnn} by learning a bank of pairwise filters.
The experiments were performed in the Q-FIRE and CASIA cross-sensor databases, reporting promising results with \gls*{eer} of $0.15$\% and  $0.31$\%, respectively.

Another deep learning application for cross-sensor iris recognition, designated DeepIrisNet, was proposed by Gangwar \& Joshi~\cite{Gangwar2016}.
In their study, two \gls*{cnn} architectures were presented and used to extract features and representations of iris images.
Comparing to the baselines, their methodology showed better robustness with respect to five different factors: effect of segmentation, image rotation, input size, training size, and network size.

Nguyen et al.~\cite{Nguyen2018} argued that generic descriptors yielding from deep learning frameworks can appropriately represent iris features from \gls{nir} images obtained in controlled environments.
The authors compared five \gls*{cnn} architectures trained in the ImageNet database~\cite{Imagenet2009}: AlexNet, VGG, Inception, ResNet and DenseNet.
Deep representations were extracted from normalized iris images at different depths of each \gls{cnn} model.
Afterward, a simple multi-class \gls*{svm} was applied to perform the identification.
The experiments were carried out in the LG2200 (ND-CrossSensor-Iris-2013) and CASIA-Iris-Thousand databases and compared with a baseline feature descriptor~\cite{Daugman2004}.
As main result, the authors argued that features extracted from intermediate layers of the networks reported better results than the representations in the deeper layers. 

Luz et al.~\cite{Luz2018} extracted deep representations of the periocular region using the VGG16 \gls{cnn} model.
The authors reported promising results by using transfer learning techniques from the face recognition domain, followed by fine-tuning using the ocular images.
The experiments achieved the state-of-the-art in the NICE.II and MobBIO databases, which were obtained in uncontrolled environments at the \gls{vis} wavelength.

Also using the NICE.II database, Silva et al.~\cite{Silva2018} proposed a fusion method of iris and periocular deep representations by means of feature selection using the Particle Swarm Optimization (PSO).
Similar to the methodology proposed in~\cite{Luz2018}, the iris and periocular deep representations were extracted with the VGG16 model trained for face recognition and fine-tuned for each trait.
Promising results were reported in the verification mode only using iris information and also using iris and periocular fusion.

Proen\c{c}a and Neves~\cite{Proenca2018} argue that periocular recognition performance is optimized when the iris and sclera regions are discarded. Also, these authors describe a processing chain based on \gls{cnn} that defines the regions-of-interest in the input image.
In their approach, a segmentation process is only required to create the training samples.
This process consists in generating a periocular image of a subject containing an ocular (sclera and iris) region belonging to other subjects.
Then, the generated samples are used for data augmentation and to feed the learning phase of the \gls{cnn} model.
The experiments were performed in the UBIRIS.v2 and FRGC databases and consistently advances the state-of-the-art in the closed-world setting. 

Zanlorensi et al.~\cite{Zanlorensi2018} evaluated the impact of the segmentation for noise removal and normalization when deep representations were extracted from the iris images.
The experiments reported that deep representations extracted from an iris bounding box without segmentation process achieved better results than normalized and segmented images.
In addition, the authors compared representations extracted from the VGG16 and ResNet50 models and the impact of using data augmentation techniques.
A new state-of-the-art was reached in the NICE.II database using only information from the iris region.

In terms of cross-sensor iris recognition, the methodology proposed by Nalla and Kumar~\cite{Nalla2017} introduced a domain adaptation framework to address this problem and reported a new approach using Markov random fields.
The experiments were performed using two cross-sensor iris databases: IIT-D CLI and ND Cross sensor 2012; and one cross-spectral iris database: \polyu.
The results reported in \polyu database in the verification protocol at closed-world achieved an \gls{eer} value of $3.97$\% in \gls{nir} \textit{vs} \gls{nir} comparisons and $6.56$\% in \gls{vis} \textit{vs} \gls{vis} comparisons.
Using the Markov random fields on cross-spectral comparisons, their methodology achieved $23.87$\% of \gls{eer}.

In~\cite{Wang2019}, the authors evaluated a range of deep learning architectures applied to the cross-spectral iris recognition.
The experimental results were performed in the \polyu and \crosseyed databases.
Experimental analysis indicates that iris features extracted from \gls{cnn} models are generally sparse and can be used for template compression. 
Several hashing algorithms were evaluated and the most effective was supervised discrete hashing achieving more accurate performance and reducing the size of iris template. 
The best results reported were achieved by incorporating supervised discrete hashing on the deep representations extracted with a \gls{cnn} model trained with a softmax cross-entropy loss.
This methodology reached an \gls{eer} value of $12.41\%$ and $6.34\%$ on the \polyu and \crosseyed databases, respectively.However, the authors do not report the system performance on the open-world protocol, which is a more realistic scenario.
Also, this methodology requires an approach for the segmentation and normalization of the iris.
To the best of our knowledge, this work is the state-of-the-art on cross-spectral recognition in the verification mode. Thus, it is used for comparison with the methodology presented in this paper.

Also applied in the cross-spectral scenario, Hernandez-Diaz et al.~\cite{Diaz2019} proposed a method using a ResNet-101 model pretrained in the Imagenet database~\cite{Imagenet2009} to extract deep representation from periocular images.
The experiments were carried out in verification mode using the IIITD Multispectral Periocular database~\cite{Sharma2014} in three different spectra: Visible, Night Vision, and  Near-Infrared.
The results were reported using features extracted at each layer from the model using chi-square distance and cosine similitude to perform the matching.
The authors stated that the features extracted from the intermediate layer from the ResNet-101 model achieved the best results in the cross-spectral experiments.

Recently, two contests were performed using the \crosseyed database, aiming to recognize iris and periocular (without the iris region) traits in a cross-spectral environment~\cite{Sequeira2016, Sequeira2017}.
However, as stated by Wang and Kumar~\cite{Wang2019}, the results reported in these competitions should be considered preliminary, as they employed a comparison protocol with less matching challenge than usual (only 3 images of each class were used in the inter-class comparison instead of all against all) and did not provide information regarding which images of each class were used in the inter-class matching (the  authors' work only reported that the images were randomly selected).
Other problems include the availability of codes and also details of the methodologies, which limit the reproducibility.

Previous works on cross-spectral recognition such as~\cite{Nalla2017, Wang2019} only use iris traits and require a methodology for iris segmentation and normalization.
Our proposal in this article combine information from the iris and periocular regions.
Also, for the iris trait, we use only a bounding box, which does not require segmentation for noise removal and normalization steps.

For completeness, there are several other applications with ocular images based on deep learning such as: spoofing detection~\cite{Menotti2015}, recognition of mislabeled left and right iris images~\cite{Du2016}, liveness detection~\cite{He2016b}, iris/periocular region location/detection~\cite{severo2018benchmark, lucio2019simultaneous}, sclera and iris segmentation~\cite{lucio2018fully, bezerra2018robust}, gender classification~\cite{Tapia2017} and sensor model identification~\cite{Marra2017}.

%% file: text/methodology.tex
\section{Methodology}
\label{sec:methodology}

\begin{figure}[!htb]
\centering
\begin{tabular}{cccc}
	{\includegraphics[width=0.18\columnwidth]{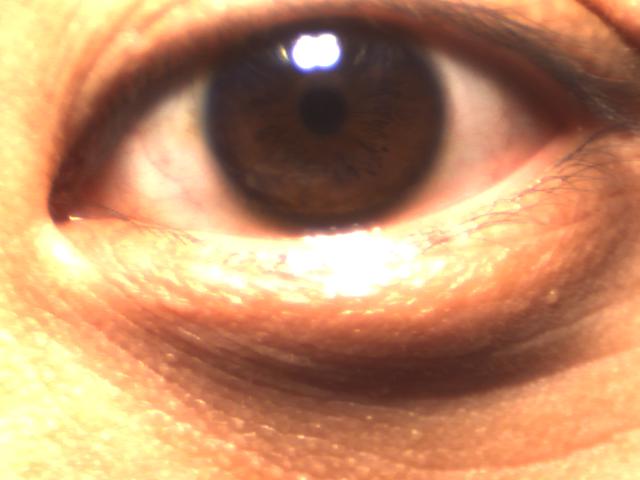}}
    {\includegraphics[width=0.18\columnwidth]{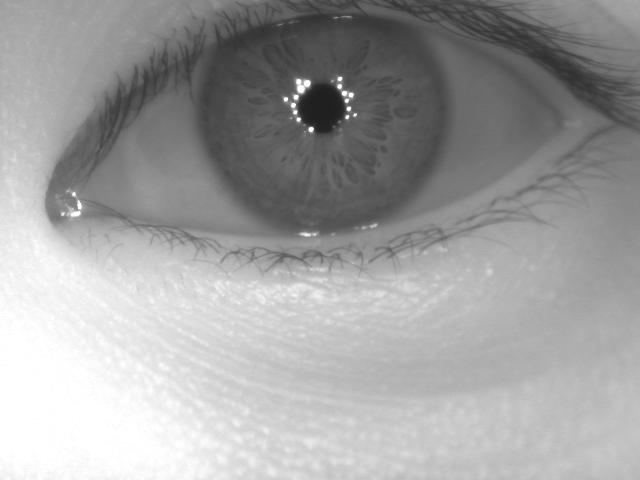}} 
    {\includegraphics[width=0.18\columnwidth]{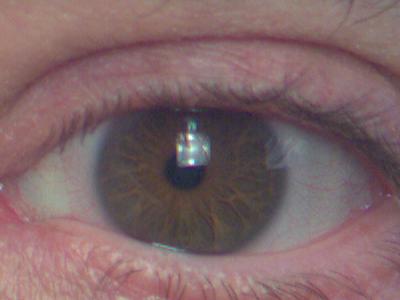}} 
    {\includegraphics[width=0.18\columnwidth]{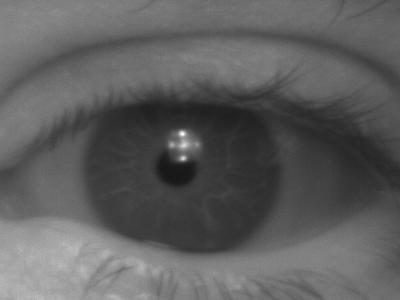}}\\ 
    
    {\includegraphics[width=0.18\columnwidth]{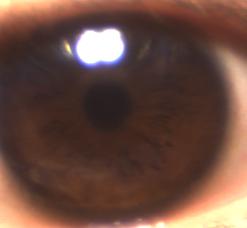}}
    {\includegraphics[width=0.18\columnwidth]{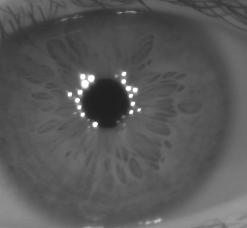}} 
    {\includegraphics[width=0.18\columnwidth]{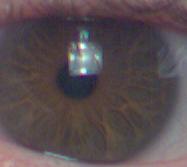}} 
    {\includegraphics[width=0.18\columnwidth]{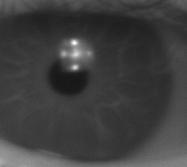}} \\

    \footnotesize
    \parbox{.18\linewidth}{\centering (a)} 
	\footnotesize
	\parbox{.18\linewidth}{ \centering (b)}
    \footnotesize
	\parbox{.18\linewidth}{ \centering (c)}
	\footnotesize
	\parbox{.18\linewidth}{ \centering (d)} \\

\end{tabular}
\caption{VIS (a,c) and NIR (b,d) samples from the \polyu (a,b) and \crosseyed (c,d) databases. First and second rows show periocular and iris images, respectively.}
\label{fig:datasamples}
\end{figure}

In this paper we analyze the use of deep representations from the eye regions (iris and periocular) on cross-spectral scenario, i.e., obtaining models able to match \gls{vis} against \gls{nir} wavelength images. Particularly, we evaluate and combine deep representations extracted from two modalities (traits): the iris and periocular regions.
In the periocular modality, features were extracted from the entire image (considering the iris, sclera, skin, eyelids and eyelashes components).
On the other way, the iris features were extracted from a bounding box, i.e., a cropped image that contains only the iris region, as described by Zanlorensi et al.~\cite{Zanlorensi2018}.
These bounding boxes were generated manually by coarse annotations and are publicly available to the research community\footnote{\supplementary}and appears in~\cite{lucio2019simultaneous}. 
Samples of the periocular and iris images used in this work are shown in Figure~\ref{fig:datasamples}.

\begin{figure}[!ht]
\begin{center}
 \includegraphics[width=.95\linewidth]{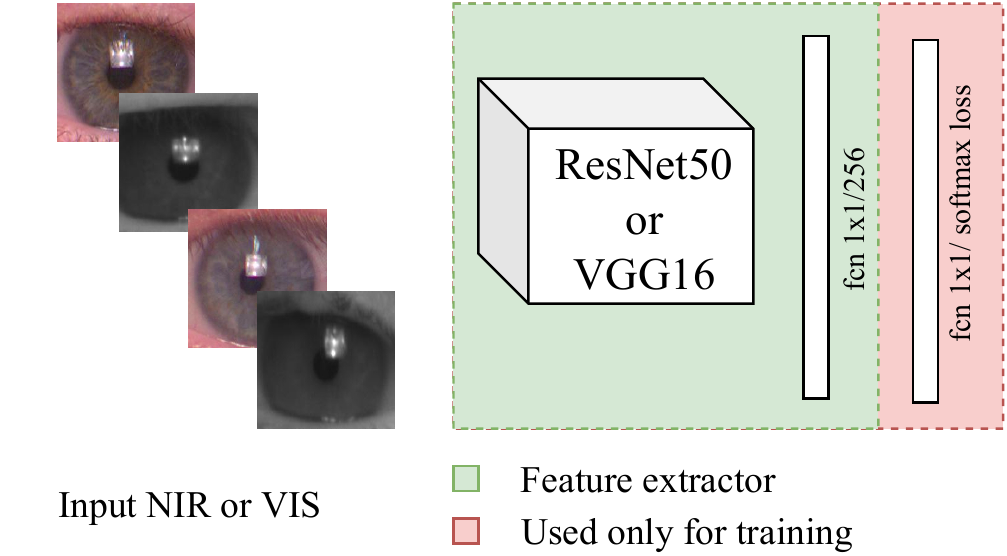}
\end{center}  
\vskip-6pt
\caption{The cross-/intra-spectral ocular recognition strategy. A single model (ResNet50 or VGG16) is used to learn features from both spectra: NIR and VIS.}
\label{fig:cnnextract}
\end{figure}

Deep representations from the periocular and iris regions were extracted using a similar approach proposed in~\cite{Zanlorensi2018}.
In this way, {the} VGG16~\cite{Omkar2015} and ResNet-50~\cite{Cao2017} \gls*{cnn} models trained for face recognition were fine-tuned to each modality.
We choose these models because they reported promising results in recent works applied in ocular recognition~\cite{Luz2018, Zanlorensi2018, Silva2018, Wang2019}.
The architecture modifications for both models consist of the removal of the last layer and the addition of two new layers.
The first one is a fully-connected layer with $256$ neurons that will be used as the feature representation and aim to reduce the feature dimensionality, since originally VGG16 and ResNet-50 have $4096$ and $2048$ features/outputs, respectively.
The other layer added has a softmax cross-entropy loss function and it is used only in the training phase in an identification mode.
We chose a feature vector of 256 features based on the results reported by Luz et al.~\cite{Luz2018}, where the authors evaluated different feature vector sizes and stated that vector with such size (256) showed the best trade-off regarding matching time, amount of memory required and matching effectiveness.
The strategy applied to extract features from \gls{nir} and \gls{vis} images is detailed in Figure~\ref{fig:cnnextract}.

The number of epochs used for training was chosen based on a validation subset composed of $20\%$ of the training set images.
After defining the number of epochs, the \gls*{cnn} models were trained using the entire training set.
The training was performed with the Stochastic Gradient Descent (SGD) optimizer and without freezing any weights of the pre-trained layers.

In the test phase, as previously mentioned, the last layer of each model was removed and the features were extracted from the first new last layer, composed by $256$ neurons.

The all-against-all matching was performed using the cosine distance metric, which measures the cosine of the angle between two vectors.
Regarding the similarity of biometrics features/representations, it is known that orientation is more important than the magnitude coefficient. 
The cosine distance metric faithfully matches this feature, being given by:
\begin{equation}
d_{c}(A,B) = 1 - \frac{\sum_{j=1}^{N}A_{j}B_{j}}{\sqrt{\sum_{j=1}^{N}A_{j}^2} \sqrt{\sum_{j=1}^{N}B_{j}^2}} \,,
\end{equation}
\noindent where $A$ and $B$ stand for the feature vectors.

The iris and periocular region representations were combined, applying the score-level fusion technique. 
Similar to approaches that also used score-level fusion for iris and periocular region traits~\cite{Ahmed2016, Ahmed2017, Nalla2017} and also based on the individual performance of each trait in our experiments, we chose to use weights of $0.6$ and $0.4$ for the periocular region and iris representations, respectively.
To perform fusion at the score-level, first, we compute the matching for each trait independently, and then we calculated the weighted arithmetic mean between the cosine distances computed for the iris and periocular modalities.

It is important to note that, in the model learning process, 
all images (\gls{nir} and \gls{vis}) were used to feed the \gls{cnn} models, making a single model to learn discriminant features of images captured in both spectra.
To the best of our knowledge, this procedure is similar to the adopted in~\cite{Wang2019} for the CNN architecture.
In the test phase the features are extracted for all images \gls{nir} or \gls{vis} images.
However, note that for evaluating the cross-spectral scenario, only images acquired under different wavelengths are paired to match.

%% file: text/protocol.tex
\section{Databases, Metrics and Protocol}
\label{sec:protocol}

This section describes the databases used, the experimental protocol defined and the metrics considered appropriate to provide a meaningful comparison between our method and the baselines.

\subsection{Databases}

Two well-known databases were used in our empirical evaluation: 1) the \polyu; and the 2) \crosseyed databases, described below:

\subsubsection{\polyu database}

\polyu (PolyU Bi-spectral) database is composed of images obtained simultaneously under both \gls{nir} and \gls{vis} wavelengths.
The entire database has $12{,}540$ images with a resolution of $640 \times 480$ pixels.
For every spectrum, there are $15$ samples of each eye (left and right) from $209$ subjects ($418$ classes)~\cite{Nalla2017}.

\subsubsection{\crosseyed database}

The \crosseyed (Cross-eyed-cross-spectral) iris database has $3{,}840$ images from $120$ subjects ($240$ classes).
There are $8$ samples from each of the classes for every spectrum.
The resolution of the images is $400 \times 300$ pixels.
All images were obtained at a distance of 1.5 meters, in an uncontrolled indoor environment, with a wide variation of ethnicity and eye colors, and lightning reflexes~\cite{Sequeira2016}.

\subsection{Metrics}

For evaluating the algorithms, \textbf{we choose the \gls*{eer} metric}, which is determined by the intersection point of \gls*{far} and \gls*{frr} curves generated when the acceptation/rejection threshold is varied.

We also report the decidability score $d'$~\cite{Daugman2003}.
The metric or index~$d'$ measures how well separated are the two types of distributions (\emph{genuines} and \emph{impostors}), in the sense that recognition errors correspond to the regions where both distributions overlap:

\begin{equation}
d' = \frac{|\mu_{E} - \mu_{I}|}{ \sqrt{ \frac{1}{2} (\sigma^{2}_{E} + \sigma^{2}_{I})} }  \, ,
\end{equation}

\noindent 
where the means and standard deviations of the genuine and impostor distributions are given by $\mu_{I}$, $\mu_{E}$, $\sigma_{I}$,  and $\sigma_{E}$, respectively.

Whereas the index~$d'$ can be related to the feature vector discrimination ability of an approach, the \gls{eer} metric measures the real performance of a biometric system.
Therefore, regarding a real-world application, we consider the \gls{eer} as the primary metric in the results reported in this work.

\subsection{Protocol}

In all experiments, the \emph{verification} setting was the unique considered, in which pairs of images are compared in order to determine whether a subject is who he claims to be or not.
For this, following a \emph{one-against-all} pairwise matching strategy, all pairs of genuine and impostor comparisons were generated.

For a fair comparison with the state-of-the-art methods, the test protocol used in this work follows the procedures given in\cite{Nalla2017, Wang2019}, which consists of a \textit{closed-world} protocol, where different instances of the same class are distributed in the training and test sets.
In the \polyu database, the first ten instances from every subject were used for training and the remainder~(five) were employed for the matching.
In the \crosseyed database, the first five instances from every subject are used for training and the remaining three instances were employed for the matching.

To perform the experiments, we considered that in both databases, the \gls{nir} and \gls{vis} images were obtained synchronously. 
Thus, here in the intra-class comparison in the cross-spectral scenario, images of the same index were not matched, because the pair represents the same image but in different spectra.
Note that in the work by Wang and Kumar~\cite{Wang2019}, the authors considered that in the \crosseyed database, non-synchronously spectrum images were obtained (based on the numbers of intra- and inter-class comparisons), so they matched \gls{nir} against \gls{vis} images of the same index in the intra-class comparison.
Then for a fair comparison with the state-of-the-art method~\cite{Wang2019}, in the closed-world protocol, we also report results considering that the \gls{nir} and \gls{vis} images where obtained non-synchronously in the \crosseyed database.

In order to evaluate the robustness of the proposed methodology, we also evaluate and then report results on the \textit{open-world} protocol, in which the training and test sets have images from different classes.
In other words, there are no images from the same subject in the training and testing.
In this protocol, for both databases, we use the first half of the subject images for training and the second half for testing.

The distributions of images and classes in the training and test sets, as well as the number of genuine and impostors pairs generated in the test phase for both databases and protocols are detailed in Table~\ref{tab:protocol}.

\begin{table}[!ht]
\centering
\caption{Genuine and impostor matches for the Closed-world (CW) and Open-world (OW) protocols on Cross- and Intra-spectral scenarios. *The comparison with the state-of-the-art methods was performed using the closed-world protocol.}
%\vspace{0.5mm}
\label{tab:protocol}
\resizebox{1\columnwidth}{!}{
\begin{tabular}{@{}lcccc@{}}
\toprule
\centering{Database}  & Protocol      & Scenario       & Train/Test Images(Classes) & Gen./Imp. pairs\\
\midrule
\polyu                & CW            & Cross & $8{,}360(418)/4{,}180(418)$     & $4{,}180/4{,}357{,}650$ \\
\polyu                & CW            & Intra & $8{,}360(418)/2{,}090(418)$     & $4{,}180/2{,}178{,}825$ \\
\polyu                & OW            & Cross & $6{,}270(209)/6{,}270(209)$     & $21{,}945/9{,}781{,}200$ \\
\polyu                & OW            & Intra  & $6{,}270(209)/3{,}135(209)$     & $21{,}945/4{,}890{,}600$\\

\crosseyed            & CW            & Cross & $2{,}400(240)/1{,}440(240)$     & $720/516{,}240$ \\
\crosseyed            & CW            & Intra  & $2{,}400(240)/720(240)$      & $720/258{,}120$  \\
\crosseyed            & OW            & Cross & $1{,}920(120)/1{,}920(120)$     & $3{,}360/913{,}920$ \\
\crosseyed            & OW            & Intra & $1{,}920(120)/960(120)$      & $3{,}360/456{,}960$ \\

\bottomrule
\end{tabular}}
\end{table}

The mean and standard deviation of $30$ repetitions for the \gls{eer} and decidability figures obtained by the proposed methodology are shown. 

%% file: text/results.tex
\section{Results and Discussion}
\label{sec:results}

In this section, we present and discuss the results observed for the intra-spectral cross-spectral scenarios, in both the iris and periocular modalities.  
We start by providing the results using the closed-world protocol, in order to establish a baseline with respect to the state-of-the-art.
We also investigate the impact of the feature vector size and the weights used to merge information from the periocular region and iris traits.
Then, the results using the open-world protocol are presented, to perceive how robust deep representations can be obtained.
Using the ResNet-50 model, a comparison of the verification effectiveness using features extracted from various network depths is performed.
Lastly, we performed a subjective analysis of the pairwise errors.

In a complementary setting, we explore the advantages yielding from fusing representations of the periocular and iris traits to improve performance.
Similar to previous works~\cite{Nalla2017, Ahmed2016, Ahmed2017} that applied higher weights in the most discriminating traits, and also considering that in all our experiments the periocular region reported better results compared to the iris, we decided to use constant weights of $0.6$ and $0.4$ respectively for the periocular and iris representations when obtaining the fused score by linear combination.

The experiments performed in this work and reported here used an NVIDIA\textsuperscript{\textregistered} Titan Xp GPU with 12GB memory and $3,840$ CUDA cores, and the  tensorflow\textsuperscript{\texttrademark} and \emph{Keras} frameworks were used to implement the \gls{cnn} models.

\subsection{Closed-world protocol}
\label{sec:closed}

At first, Table~\ref{tab:PolyUclosed} and Table~\ref{tab:CrossEyedclosed} report the results observed for verification mode, in the cross-spectral and intra-spectral scenarios (\gls{nir} against \gls{nir} and \gls{vis} against \gls{vis}) and using the closed-world protocol.
In a way similar to Nalla and Kumar~\cite{Nalla2017} and also to guarantee a fair comparison to their method, the fusion of two spectra on the \polyu database was carried out by linear combination, using weights of $0.6$ and $0.4$, respectively, to the \gls{nir} and \gls{vis} images.
However, based on the individual spectral results, on the \crosseyed database, we used weights of $0.6$ and $0.4$ for the \gls{vis} and \gls{nir} representations, respectively.
Also, on the \crosseyed database, we can perceive that the spectral fusion using iris representations extracted by the VGG16 model reported lower results than using the only \gls{vis} spectral information. The results show that the representations obtained from \gls{nir} images presented a high \gls{eer} value, which penalized the fusion of spectra. Therefore, lower weight for \gls{nir} representations may improve the fusion result.
The results of those fusions are shown in Table~\ref{tab:PolyUclosed} and Table~\ref{tab:CrossEyedclosed} (VIS and NIR Fusion section).

\begin{table}[!ht]
\centering
\caption{Results - closed-world protocol on the \polyu database. *Using only $140$ subjects from a total of $209$.}
\label{tab:PolyUclosed}
\resizebox{\columnwidth}{!}
{
\begin{tabular}{@{}lccc@{}}
\toprule
\centering{Approach}              & Modality   & {EER (\%)}        & {Decidability} \\
\midrule
\multicolumn{4}{c}{Cross-Spectral} \\
\midrule
CNN with SDH~\cite{Wang2019}*       & Iris       & $5.39$            & $2.13$ \\
CNN with SDH~\cite{Wang2019}        & Iris       & $12.41$           & $-$ \\
VGG16 with SDH~\cite{Wang2019}*     & Iris       & $4.85$            & $-$ \\
Proposed VGG16                      & Iris       & $2.16\pm0.16$     & $5.23\pm0.08$ \\
ResNet50 with SDH~\cite{Wang2019}*  & Iris       & $7.17$            & $-$ \\
\textbf{Proposed ResNet50}          & \textbf{Iris} & \boldmath{$1.13\pm0.14$} & \boldmath{$5.17\pm0.08$} \\

Proposed VGG16                      & Periocular & $1.80\pm0.21$     & $6.03\pm0.20$ \\
\textbf{Proposed ResNet50}                   & \textbf{Periocular} & \boldmath{$0.78\pm0.09$}     & \boldmath{$5.97\pm0.08$} \\

Proposed VGG16                      & Fusion     & $0.93\pm0.10$     & $6.97\pm0.13$ \\
\textbf{Proposed ResNet50}          & \textbf{Fusion}     & \boldmath{$0.49\pm0.06$} & \boldmath{$6.75\pm0.08$} \\
\midrule
\multicolumn{4}{c}{VIS vs VIS} \\
\midrule
Nalla and Kumar~\cite{Nalla2017}*                 & Iris       & $6.56$            & $-$ \\
Proposed VGG16                    & Iris       & $1.53\pm0.12$     & $6.27\pm0.08$ \\
\textbf{Proposed ResNet50}        & \textbf{Iris}       & \boldmath{$0.78\pm0.08$}     & \boldmath{$5.91\pm0.07$} \\
Proposed VGG16                    & Periocular & $1.50\pm0.16$     & $6.63\pm0.21$ \\
\textbf{Proposed ResNet50}        & \textbf{Periocular} & \boldmath{$0.61\pm0.11$}     & \boldmath{$6.57\pm0.08$} \\
Proposed VGG16                    & Fusion     & $0.76\pm0.10$     & $7.73\pm0.14$ \\
\textbf{Proposed ResNet50}        & \textbf{Fusion}     & \boldmath{$0.35\pm0.06$}     & \boldmath{$7.44\pm0.10$} \\
\midrule
\multicolumn{4}{c}{NIR vs NIR} \\ 
\midrule
Nalla and Kumar~\cite{Nalla2017}*                 & Iris       & $3.97$            & $-$ \\
Proposed VGG16                    & Iris       & $1.21\pm0.13$     & $6.61\pm0.10$ \\
\textbf{Proposed ResNet50}        & \textbf{Iris}       & \boldmath{$0.68\pm0.07$}     & \boldmath{$6.05\pm0.07$} \\
Proposed VGG16                    & Periocular & $1.56\pm0.19$     & $6.58\pm0.21$ \\
\textbf{Proposed ResNet50}        & \textbf{Periocular} & \boldmath{$0.68\pm0.10$}     & \boldmath{$6.59\pm0.07$} \\
Proposed VGG16                    & Fusion     & $0.70\pm0.11$     & $7.86\pm0.17$ \\
\textbf{Proposed ResNet50}        & \textbf{Fusion}     & \boldmath{$0.40\pm0.06$}     & \boldmath{$7.54\pm0.09$} \\
\midrule
\multicolumn{4}{c}{VIS and NIR Fusion} \\ 
\midrule
Nalla and Kumar~\cite{Nalla2017}*                 & Iris       & $2.86$            & $-$ \\
Proposed VGG16                    & Iris       & $1.01\pm0.09$     & $6.81\pm0.08$ \\
\textbf{Proposed ResNet50}        & \textbf{Iris}       & \boldmath{$0.59\pm0.08$}     & \boldmath{$6.29\pm0.07$} \\
Proposed VGG16                    & Periocular & $1.36\pm0.15$     & $6.79\pm0.21$ \\
\textbf{Proposed ResNet50}         & \textbf{Periocular} & \boldmath{$0.56\pm0.10$}     & \boldmath{$6.82\pm0.08$} \\
Proposed VGG16                    & Fusion     & $0.63\pm0.10$     & $8.05\pm0.16$ \\
\textbf{Proposed ResNet50}                 & \textbf{Fusion}     & \boldmath{$0.35\pm0.05$}     & \boldmath{$7.75\pm0.10$} \\

\bottomrule
\end{tabular}
}
\end{table}

\begin{table}[!ht]
\centering
\caption{Results - closed-world protocol on the \crosseyed database. *same protocol used by Wang and Kumar~\cite{Wang2019}.}
\label{tab:CrossEyedclosed}
\resizebox{\columnwidth}{!}{
\begin{tabular}{@{}lccc@{}}
\toprule
\centering{Approach}               & Modality   & {EER (\%)}        & {Decidability} \\
\midrule
\multicolumn{4}{c}{Cross-spectral} \\
\midrule
CNN with SDH~\cite{Wang2019}      & Iris                  & $6.34$                         & $2.54$ \\
VGG16 with SDH~\cite{Wang2019}    & Iris                  & $3.13$                         & $-$ \\
Proposed VGG16*                   & Iris                  & $5.58\pm0.59$                  & $3.87\pm0.16$ \\
Proposed VGG16                    & Iris                  &$6.76\pm0.56$                   &$3.58\pm0.14$ \\
ResNet50 with SDH~\cite{Wang2019} & Iris                  & $6.11$                         & $-$ \\
\textbf{Proposed ResNet50}*       & \textbf{Iris}         & \boldmath{$2.45\pm0.25$}       & \boldmath{$4.73\pm0.09$} \\
\textbf{Proposed ResNet50}        & \textbf{Iris}         & \boldmath{$3.07\pm0.38$}       & \boldmath{$4.49\pm0.09$} \\
Proposed VGG16*                   & Periocular            & $2.35\pm0.28$                  & $5.61\pm0.20$ \\
Proposed VGG16                    & Periocular            & $3.18\pm0.42$                  & $5.19\pm0.21$ \\
\textbf{Proposed ResNet50}*       & \textbf{Periocular}   & \boldmath{$1.45\pm0.24$}       & \boldmath{$4.73\pm0.09$} \\
\textbf{Proposed ResNet50}        & \textbf{Periocular}   & \boldmath{$1.95\pm0.35$}       & \boldmath{$5.34\pm0.12$} \\
Proposed VGG16*                   & Fusion                & $1.86\pm0.19$                  & $5.78\pm0.11$ \\
Proposed VGG16                    & Fusion                & $2.66\pm0.29$                  & $5.31\pm0.12$ \\
\textbf{Proposed ResNet50}*       & \textbf{Fusion}       & \boldmath{$1.06\pm0.15$}       & \boldmath{$6.29\pm0.11$} \\
\textbf{Proposed ResNet50}        & \textbf{Fusion}       & \boldmath{$1.40\pm0.26$}       & \boldmath{$5.93\pm0.12$} \\
\midrule
\multicolumn{4}{c}{VIS vs VIS} \\
\midrule
Proposed VGG16                    & Iris                  & $3.66\pm0.39$                  & $4.85\pm0.16$ \\
\textbf{Proposed ResNet50}        & \textbf{Iris}         & \boldmath{$2.47\pm0.42$}       & \boldmath{$5.12\pm0.13$} \\
Proposed VGG16                    & Periocular            & $2.60\pm0.40$                  & $5.57\pm0.21$ \\
\textbf{Proposed ResNet50}        & \textbf{Periocular}   & \boldmath{$1.70\pm0.37$}       & \boldmath{$5.66\pm0.13$} \\
Proposed VGG16                    & Fusion                & $1.94\pm0.29$                  & $6.15\pm0.16$ \\
\textbf{Proposed ResNet50}        & \textbf{Fusion}       & \boldmath{$1.17\pm0.25$}       & \boldmath{$6.39\pm0.13$} \\
\midrule
\multicolumn{4}{c}{NIR vs NIR} \\ 
\midrule
Proposed VGG16                    & Iris                  & $7.31\pm0.91$                  & $3.46\pm0.18$ \\
\textbf{Proposed ResNet50}        & \textbf{Iris}         & \boldmath{$2.74\pm0.34$}       & \boldmath{$4.72\pm0.08$} \\
Proposed VGG16                    & Periocular            & $2.97\pm0.46$                  & $5.36\pm0.23$ \\
Proposed ResNet50                 & \textbf{Periocular}   & \boldmath{$1.78\pm0.39$}       & \boldmath{$5.54\pm0.13$} \\
Proposed VGG16                    & Fusion                & $2.40\pm0.35$                  & $5.36\pm0.12$ \\
\textbf{Proposed ResNet50}        & \textbf{Fusion}       & \boldmath{$1.31\pm0.24$}       & \boldmath{$6.14\pm0.12$} \\
\midrule
\multicolumn{4}{c}{VIS and NIR Fusion} \\ 
\midrule
Proposed VGG16                    & Iris                  & $3.69\pm0.39$                  & $4.65\pm0.15$ \\ 
\textbf{Proposed ResNet50}        & \textbf{Iris}         & \boldmath{$2.18\pm0.31$}       & \boldmath{$5.25\pm0.10$} \\ 
Proposed VGG16                    & Periocular            & $2.44\pm0.43$                  & $5.70\pm0.22$ \\ 
\textbf{Proposed ResNet50}        & \textbf{Periocular}   & \boldmath{$1.54\pm0.30$}       & \boldmath{$5.76\pm0.13$} \\ 
Proposed VGG16                    & Fusion                & $1.92\pm0.29$                  & $6.09\pm0.14$ \\ 
\textbf{Proposed ResNet50}        & \textbf{Fusion}       & \boldmath{$1.11\pm0.20$}       & \boldmath{$6.47\pm0.12$} \\ 

\bottomrule
\end{tabular}}
\end{table}

Anyway, it can be seen that - for both databases - the proposed approach achieves better results than the state-of-the-art methods, both in the cross-spectral and in the intra-spectral scenarios
even that the protocol used in this paper is more challenging.
For example, in the \polyu database, we used images from all 209 subjects in the experiments, while the approaches proposed by Wang and Kumar~\cite{Wang2019}, and Nalla and Kumar~\cite{Nalla2017} used images from only 140 subjects.
In the \crosseyed database, based on the number of pairs of intra-class comparisons reported in the experiments by Wang and Kumar~\cite{Wang2019}, the authors considered that the database has images obtained non-synchronously.
Images from the \crosseyed database were obtained using a dual sensor with a beam splitter, so the \gls{nir} and  \gls{vis} images are acquired simultaneously. 
However, we visually verified that the images of the same index, i.e., those that should be the same one in the \gls{nir} and  \gls{vis}, have a random shift in each spectrum.
Thus, for a fair comparison with the state-of-the-art approaches, we report the results using both protocols, considering the images obtained synchronously and non-synchronously.
Note that we collected the state-of-the-art results from the original papers~\cite{Nalla2017, Wang2019}, i.e., we did not have implemented any approach from these works.

In terms of the CNN architectures, the ResNet-50 model reported lower \gls{eer} values compared to the VGG16 model in all cases.
However, in some cases, specifically in the \polyu database, the representations extracted with the VGG16 model obtained a better separation of intra- and inter-class distributions, as can be seen in their Decidability index.

The results show that in \crosseyed, the periocular modality achieves better results than the iris one.
However, in the \polyu database, there is no significant difference between iris and periocular representations, mainly in the intra-spectral experiments.
From a visual inspection analysis of the pairwise comparison errors (some examples are shown in Section~\ref{ssec:subjectiveevaluation}), we perceive that in the \polyu database, some uncontrolled conditions present in the images such as pose, eye gaze, and rotation may penalize the quality of the periocular representations.
These conditions are more controlled in the cross-eyed images.
Also, \crosseyed images are smaller than \polyu images, so the iris region is even smaller, and the periocular images are better centralized based on the iris region in the \crosseyed and not in the \polyu database.
Nevertheless, \crosseyed images present a more significant difference in color and illumination among classes, which makes them more distinct and may explain the better results in \gls{vis} against \gls{vis} comparisons than \gls{nir} against \gls{nir}.

\subsection{Feature size and fusion weights analyses}
\label{sec:fusionsize}

In this section, we analyze and discuss the impact of feature vector size and the weights used for the fusion of the iris and periocular region representations.

As state in Section~\ref{sec:methodology}, we choose the feature size of $256$ based on the experiments and results reported by Luz et al~\cite{Luz2018}.
Therefore, we also performed some experiments creating new models with different sizes in the last layer before the softmax one, i.e., the layer used to extract the features (representations). 
The results of the fusion of iris and periocular representations extracted with these models are presented in Table~\ref{tab:featsize}.
Luz et al.~\cite{Luz2018} stated that for the cosine distance metric, high dimensional vectors resulted in better performance.
Conversely, our results show that representations extracted with the ResNet50 model achieve lower values of \gls{eer} when the feature vector is smaller.
The same occurs in the VGG16 model features in the \polyu database.
Regarding the decidability index, the size of the feature vector does not show to have much impact.
These results may be related to the fact that both models can generate sparse feature vectors, as stated by Wang and Kumar~\cite{Wang2019}. 
Thus a bigger feature vector will not always improve the performance of the biometric system.
Here, we decided to keep a feature vector size of 256 because it keep a trade-off between \gls{eer} and Decidability.

\begin{figure*}[!htb]
\centering
\begin{tabular}{cc}

    \parbox{\columnwidth}{\centering VGG16 Features} 
	\parbox{\columnwidth}{ \centering ResNet50 Features} \\

	\includegraphics[width=\columnwidth]{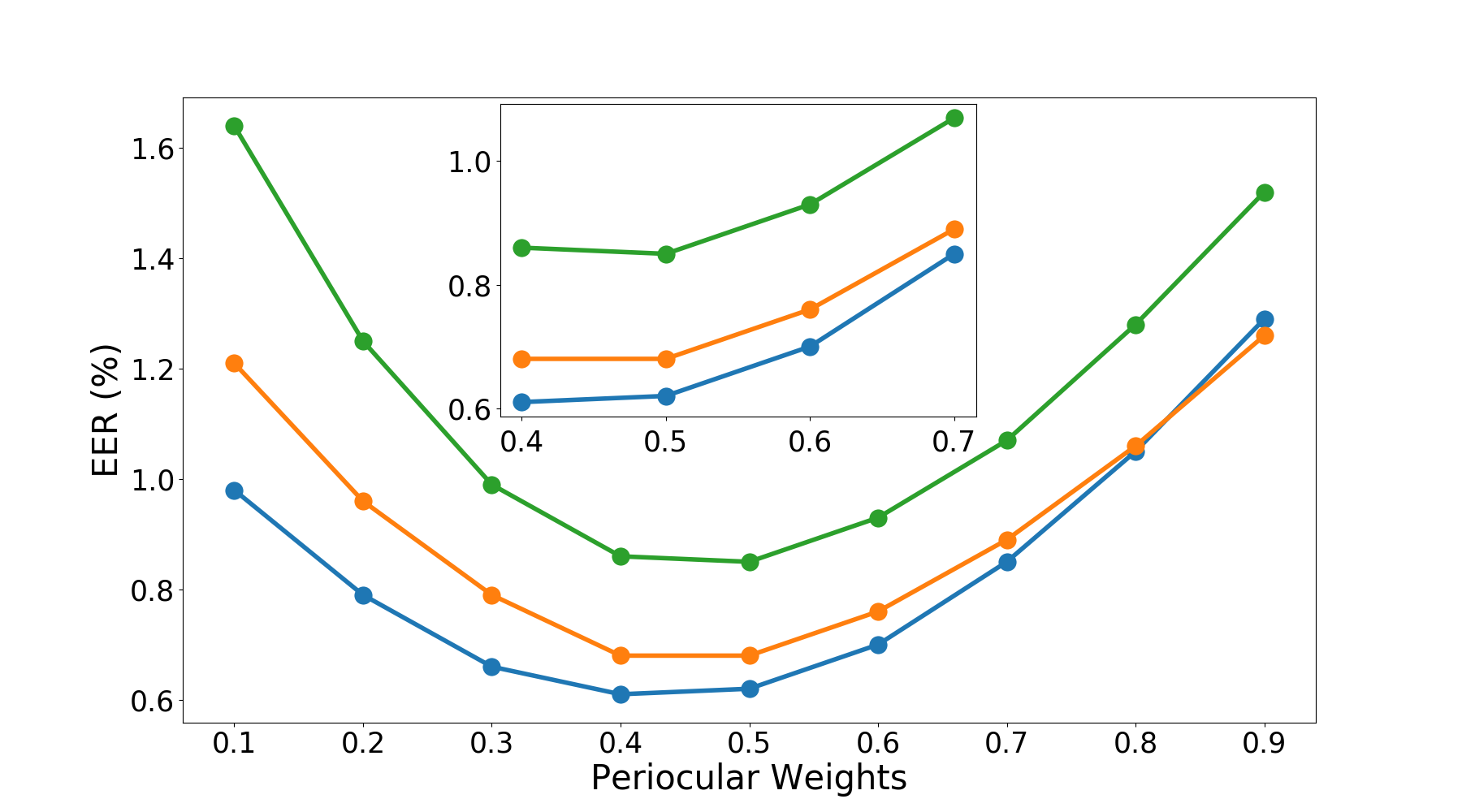}
    \includegraphics[width=\columnwidth]{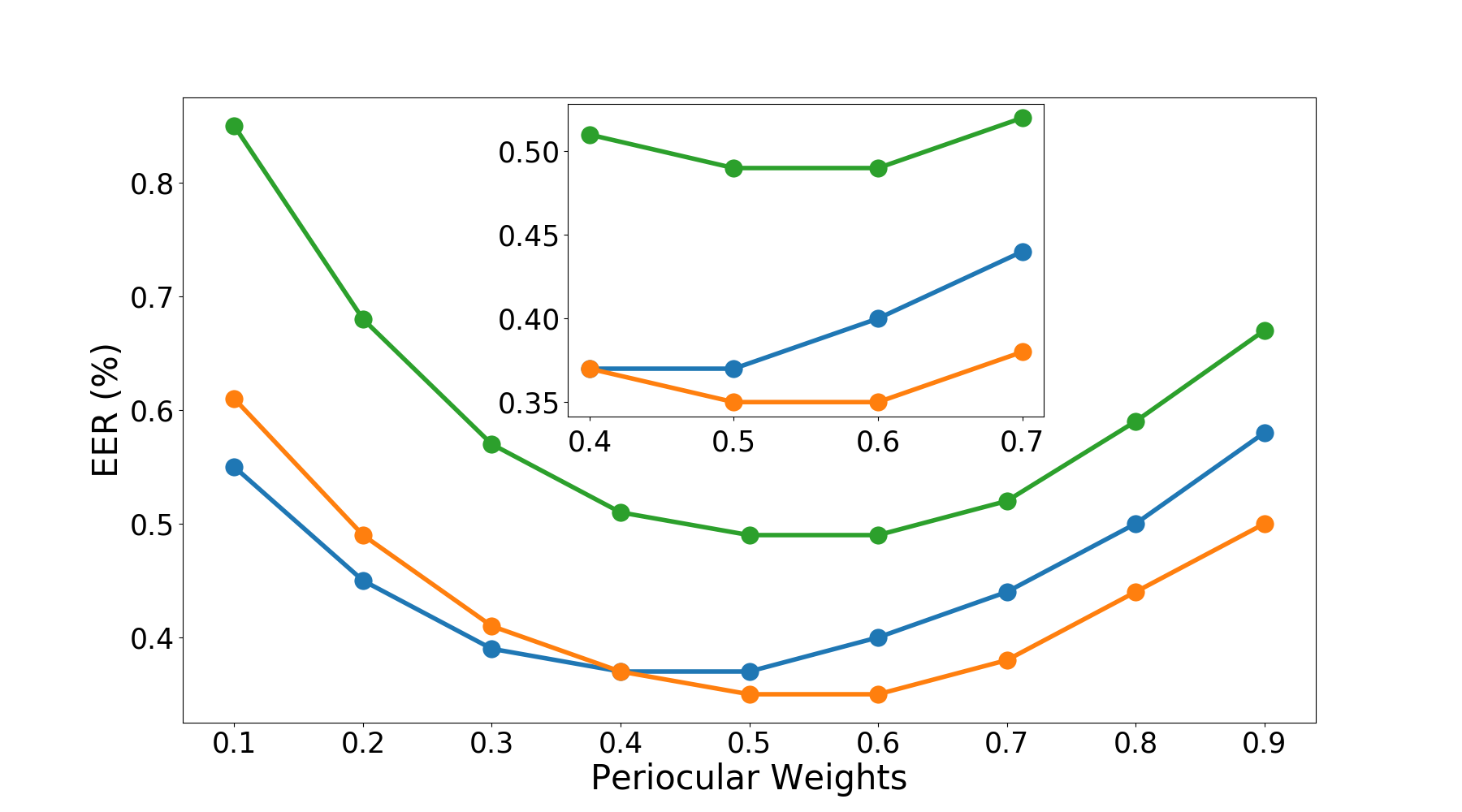}\\ 
    
    \includegraphics[width=\columnwidth]{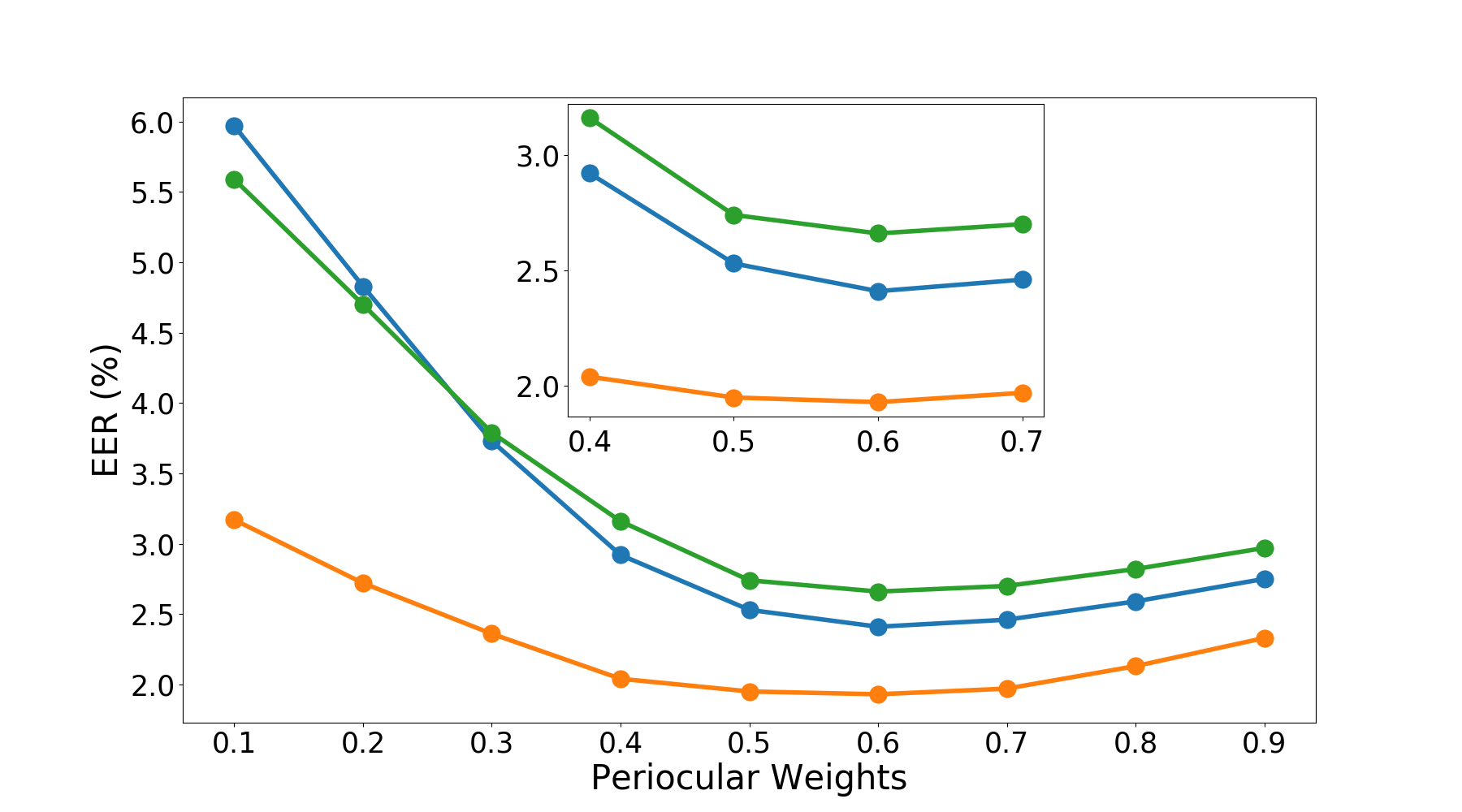}
    \includegraphics[width=\columnwidth]{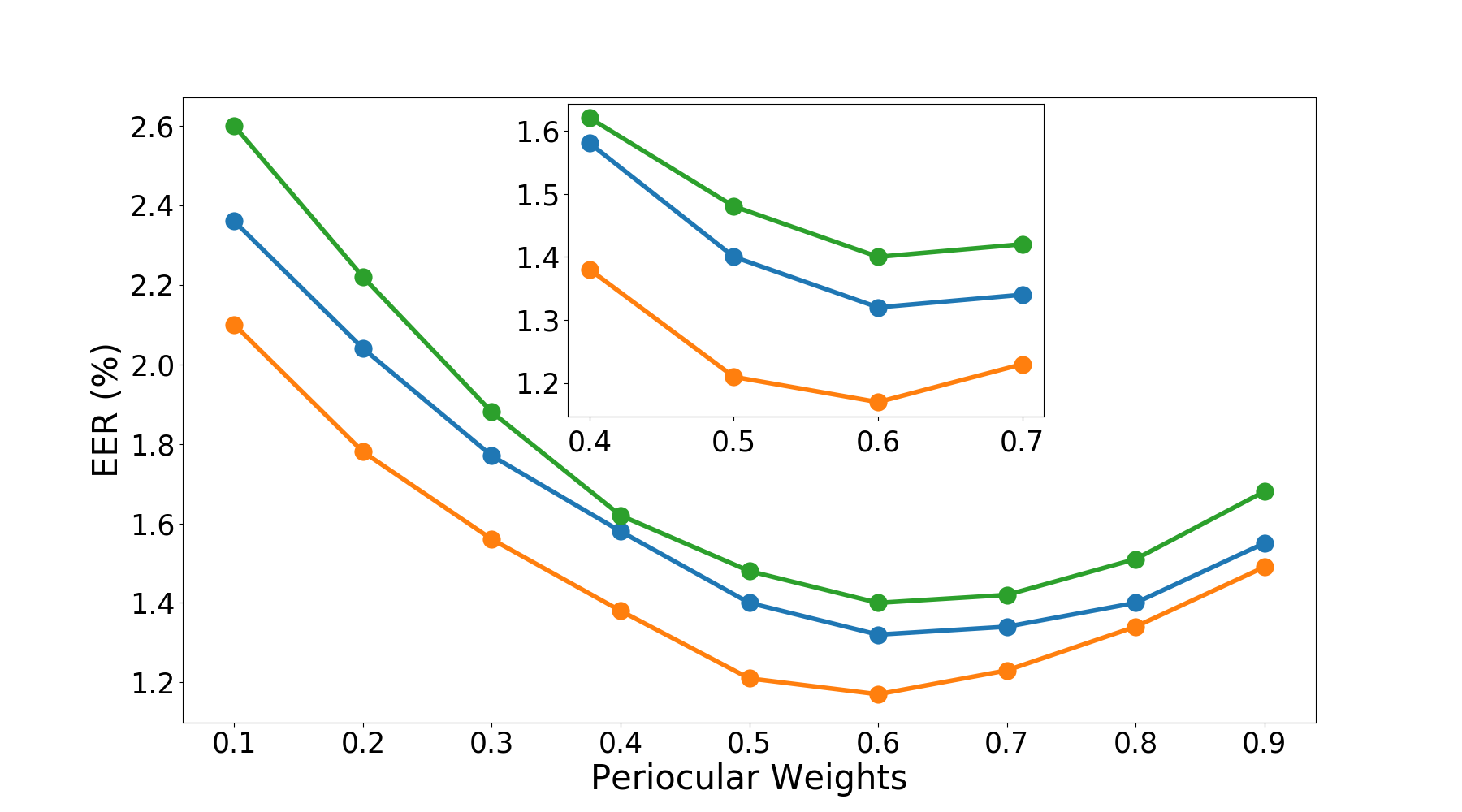} \\

    \includegraphics[width=.65\columnwidth]{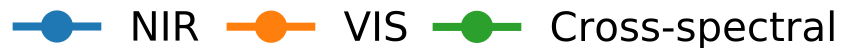}\\

\end{tabular}
\caption{Periocular weights impact on the traits fusion in the cross-spectral scenario on the \polyu (top row) and \crosseyed (bottom row) databases.}
\label{fig:weights}
\end{figure*}

\begin{table}[!ht]
\centering
\caption{Feature vector size results fusing iris and periocular region traits on Cross-spectral scenario.}
\label{tab:featsize}
\vspace{0.5mm}
\resizebox{\columnwidth}{!}{
\begin{tabular}{@{}lccccc@{}}
\toprule

\centering \multirow{2}{*}{Model} & \multirow{2}{*}{Feat. Size} & \multicolumn{2}{c}{\polyu} & \multicolumn{2}{c}{\crosseyed} \\

\cmidrule{3-6}

                       &                             & EER (\%)         & Decidability         & EER (\%)            & Decidability           \\

\midrule
\multirow{6}{*}{ResNet50}   & $1024$          & $0.54\pm0.09$  & $6.76\pm0.10$  & $1.61\pm0.25$  & $5.93\pm0.13$ \\
                            & $512$           & $0.56\pm0.06$  & $6.73\pm0.08$ & $1.35\pm0.22$  & $6.00\pm0.11$ \\
                            & $256$           & $0.49\pm0.06$  & $6.75\pm0.08$ & $1.40\pm0.26$  & $5.93\pm0.12$ \\
                            & $128$           & $0.43\pm0.05$  & $6.70\pm0.08$ & $1.35\pm0.30$  & $5.99\pm0.13$ \\
                            & $64$            & $0.37\pm0.07$  & $6.50\pm0.08$ & $1.26\pm0.22$  & $5.93\pm0.15$ \\
                            & $32$            & $0.30\pm0.05$  & $6.05\pm0.15$ & $1.41\pm0.27$  & $5.65\pm0.16$ \\
\midrule
\multirow{6}{*}{VGG16}      & $1024$          & $0.99\pm0.10$  & $6.85\pm0.08$ & $2.68\pm0.28$  & $5.29\pm0.11$ \\
                            & $512$           & $0.92\pm0.12$  & $6.94\pm0.11$ & $2.53\pm0.38$  & $5.35\pm0.14$ \\
                            & $256$           & $0.93\pm0.10$  & $6.97\pm0.13$ & $2.66\pm0.29$  & $5.31\pm0.12$ \\
                            & $128$           & $0.80\pm0.12$  & $7.03\pm0.10$ & $2.78\pm0.33$  & $5.28\pm0.10$ \\
                            & $64$            & $0.73\pm0.11$  & $6.93\pm0.11$ & $2.67\pm0.37$  & $5.23\pm0.15$ \\
                            & $32$            & $0.69\pm0.10$  & $6.46\pm0.07$ & $2.79\pm0.47$  & $4.98\pm0.17$ \\
\bottomrule
\end{tabular}}
\end{table}

As described in Section~\ref{sec:methodology}, similar to some approaches~\cite{Ahmed2016, Ahmed2017, Nalla2017} in the literature and based on the individual performance in our experiments, we choose weights of $0.6$ and $0.4$ for the periocular and iris fusion, respectively.
Nevertheless, in this section, we evaluated the impact of different iris and periocular weights on the trait representations fusion in the cross-spectral scenario, for both models.
Indeed, we impose $w_p \in [0,1]$, such that $w_i+w_p=1$, where $w_p$ and $w_i$ stand for the periocular and iris weights, respectively.
The results are reported in Figure~\ref{fig:weights}.

Even though the values of \gls{eer} are lower using features extracted with the ResNet50 model, we can observe a similar behaviour regarding the weight difference in both databases for both models.
That is, when the weights are appropriately combined the best results are achieved.
We can also observe that the periocular trait has more impact on the \crosseyed database than on the \polyu database.
We also note that on the \polyu database, in some cases, fusion with a higher iris weight ($w_i = 0.6$ and $w_p=0.4$ using VGG16 features) may achieve a lower value of \gls{eer}.

\subsection{Open-world protocol}

Also, the experimental results observed for the open-world scenario are presented in Table~\ref{tab:tab:polyuow} and Table~\ref{tab:crosseyedow} for the \polyu and \crosseyed databases, respectively.
Notice that this protocol is more challenging since there is no sample of the test classes in the training set. Another factor that makes it more difficult is that compared to the closed-world protocol, fewer images are available for model training, and there are more images on the test set increasing the pair of genuine and imposter comparisons.

\begin{table}[!ht]
\centering
\caption{Verification in the open-world protocol on the \polyu database.}
\label{tab:tab:polyuow}
\resizebox{\columnwidth}{!}{
\begin{tabular}{@{}lccc@{}}
\toprule
\centering{Approach}               & Modality   & {EER (\%)}        & {Decidability} \\
\midrule
\multicolumn{4}{c}{Cross-spectral} \\
\midrule
Proposed ResNet50                 & Iris       & $12.01\pm0.78$     & $2.44\pm0.08$ \\
Proposed ResNet50                 & Periocular & $8.02\pm0.65$      & $3.00\pm0.11$ \\
Proposed ResNet50                 & Fusion     & $6.01\pm0.39$      & $3.35\pm0.08$ \\

\midrule
\multicolumn{4}{c}{VIS vs VIS} \\
\midrule
Proposed ResNet50                  & Iris       & $4.30\pm0.24$     & $3.86\pm0.07$ \\
Proposed ResNet50                  & Periocular & $3.94\pm0.27$     & $4.14\pm0.09$ \\
Proposed ResNet50                  & Fusion     & $2.61\pm0.11$     & $4.71\pm0.06$ \\
\midrule
\multicolumn{4}{c}{NIR vs NIR} \\ 
\midrule
Proposed ResNet50                  & Iris       & $4.00\pm0.24$     & $3.88\pm0.08$ \\
Proposed ResNet50                  & Periocular & $4.00\pm0.26$     & $4.10\pm0.10$ \\
Proposed ResNet50                  & Fusion     & $2.55\pm0.17$     & $4.68\pm0.10$ \\

\bottomrule
\end{tabular}}
\end{table}

\begin{table}[!ht]
\centering
\caption{Results - open-world protocol on the \crosseyed database.}
\label{tab:crosseyedow}
\resizebox{\columnwidth}{!}{
\begin{tabular}{@{}lccc@{}}
\toprule
\centering{Approach}     & Modality       & {EER (\%)}        & {Decidability} \\
\midrule
\multicolumn{4}{c}{Cross-spectral} \\
\midrule
Proposed ResNet50        & Iris           & $8.87\pm0.77$     & $2.85\pm0.11$ \\
Proposed ResNet50        & Periocular     & $4.39\pm0.44$     & $3.85\pm0.11$ \\
Proposed ResNet50        & Fusion         & $3.51\pm0.32$     & $4.17\pm0.07$ \\
\midrule
\multicolumn{4}{c}{VIS vs VIS} \\
\midrule
Proposed ResNet50       & Iris            & $4.25\pm0.35$  & $4.01\pm0.10$ \\
Proposed ResNet50       & Periocular      & $3.41\pm0.38$  & $4.41\pm0.11$ \\
Proposed ResNet50       & Fusion          & $2.57\pm0.26$  & $4.97\pm0.09$ \\
\midrule
\multicolumn{4}{c}{NIR vs NIR} \\ 
\midrule
Proposed ResNet50       & Iris            & $5.04\pm0.43$  & $3.63\pm0.12$ \\
Proposed ResNet50       & Periocular      & $3.51\pm0.40$  & $4.38\pm0.12$ \\
Proposed ResNet50       & Fusion          & $2.75\pm0.28$  & $4.83\pm0.10$ \\

\bottomrule
\end{tabular}}
\end{table}

\begin{figure*}[!tb]
\begin{center}
 \includegraphics[width=.95\linewidth]{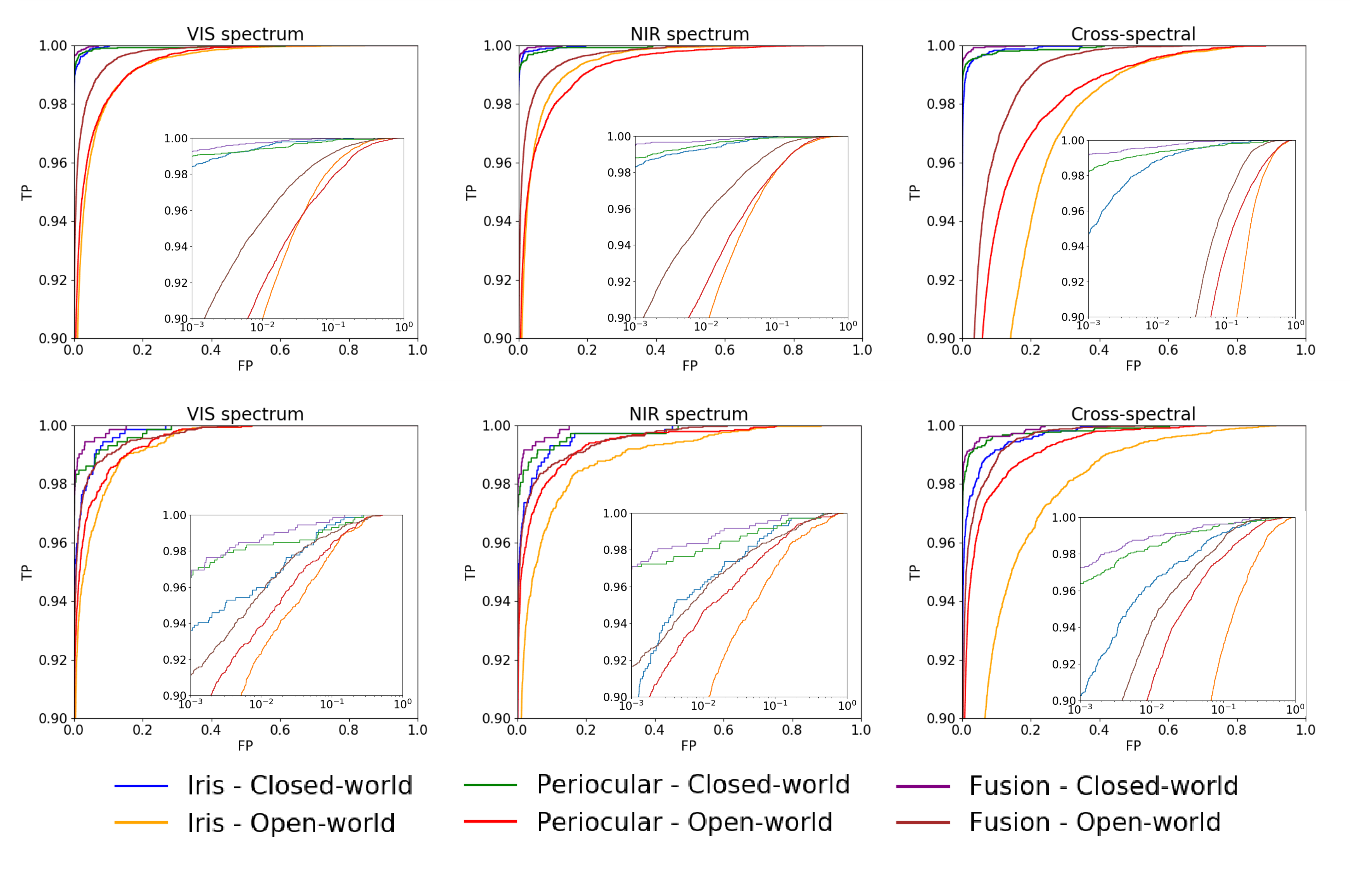}
\end{center}  
\vskip-6pt
\caption{ROC curves comparing the closed- and open-world protocols on the \polyu (top row) and \crosseyed (bottom row) databases.}
\label{fig:ow}
\end{figure*}

To perceive the differences in performance, a comparison of the results using closed- and open-world is shown with the \gls{roc} curve in Figure~\ref{fig:ow}.
Even though a fully fair comparison between closed- and open-world protocols is not feasible because the number of subjects used for learning is different, it is noticeable that the open-world protocol reported worse performance in all modes compared to the closed-world protocol.
Nevertheless, we conclude that fusing the ocular and iris representations also leads to promising results in the open-world protocol, given that the observed decidability was higher than three for both databases considered.

\subsection{ResNet-50: Performance vs. Network Depth}

Having concluded that the \emph{ResNet-50} yields to the optimal results in terms of EER in our experiments, our next goal was to perceive how the verification performance varies with respect to the depth of the layer from where representations are taken. 
In this experiment, we considered all the convolution layers with stride equal to $2$, resulting in four different depths to be tested: $12$, $24$, $42$ and $50$ layers.
%layers 37(12), 79(24), 141(42), 173(50)
For each one of the four possibilities (depths), the same modifications described in the methodology section were made, adding a fully-connected layer with $256$ neurons and a layer with a softmax cross-entropy loss function.
The verification results using the different depths are reported in Table~\ref{tab:depth} for the \polyu and \crosseyed databases.

\begin{table}[!ht]
\centering
\caption{\gls{eer} values observed for different depths (trainable parameters) of ResNet50 architecture, using the closed-world protocol.}
\label{tab:depth}
\vspace{0.5mm}
\resizebox{\columnwidth}{!}{
\begin{tabular}{@{}cccccc@{}}
\toprule
\centering \multirow{2}{*}{Spec.} & \multirow{2}{*}{Trait}  & {$12$ layers} & {$24$ layers} & {$42$ layers}  & {$50$ layers} \\
\centering                           &                            & {(26M)}       & {(14.5M)}     & {(15.6M)}       & {(24.1M)} \\

\midrule
\multicolumn{6}{c}{\polyu} \\
\midrule

VIS    & Iris       & $3.21\pm0.16$   & $2.29\pm0.15$   & $1.60\pm0.10$   & $0.78\pm0.08$ \\
VIS    & Perioc.    & $3.84\pm0.14$   & $3.17\pm0.18$   & $2.17\pm0.12$   & $0.61\pm0.11$ \\
VIS    & Fusion     & $1.66\pm0.06$   & $1.41\pm0.07$   & $1.06\pm0.11$   & $0.35\pm0.06$ \\
NIR    & Iris       & $3.55\pm0.18$   & $2.36\pm0.11$   & $1.46\pm0.10$   & $0.68\pm0.07$ \\
NIR    & Perioc.    & $4.16\pm0.17$   & $3.39\pm0.18$   & $2.27\pm0.14$   & $0.68\pm0.10$ \\
NIR    & Fusion     & $2.13\pm0.08$   & $1.56\pm0.08$   & $1.09\pm0.10$   & $0.40\pm0.06$ \\
Cross  & Iris       & $6.39\pm0.41$   & $4.50\pm0.23$   & $3.09\pm0.19$   & $1.13\pm0.14$ \\
Cross  & Perioc.    & $5.38\pm0.20$   & $4.04\pm0.17$   & $2.71\pm0.14$   & $0.78\pm0.09$ \\
Cross  & Fusion     & $2.95\pm0.15$   & $2.07\pm0.13$   & $1.41\pm0.09$   & $0.49\pm0.06$ \\

\midrule
\multicolumn{6}{c}{\crosseyed} \\
\midrule
VIS    & Iris       & $4.77\pm0.38$   & $3.29\pm0.26$   & $2.16\pm0.34$   & $2.47\pm0.42$ \\
VIS    & Perioc.    & $6.34\pm0.36$   & $3.70\pm0.35$   & $1.90\pm0.23$   & $1.70\pm0.37$ \\
VIS    & Fusion     & $3.78\pm0.22$   & $1.94\pm0.16$   & $1.25\pm0.18$   & $1.17\pm0.25$ \\
NIR    & Iris       & $20.24\pm0.70$  & $16.28\pm0.66$  & $8.78\pm0.56$   & $2.74\pm0.34$ \\
NIR    & Perioc.    & $7.28\pm0.35$   & $4.08\pm0.32$   & $1.88\pm0.23$   & $1.78\pm0.39$ \\
NIR    & Fusion     & $7.78\pm0.30$   & $4.90\pm0.33$   & $2.03\pm0.23$   & $1.31\pm0.24$ \\

Cross  & Iris       & $20.88\pm0.74$  & $15.91\pm0.60$  & $8.12\pm0.63$   & $3.07\pm0.38$ \\
Cross  & Perioc.    & $7.53\pm0.38$   & $4.17\pm0.38$   & $2.31\pm0.31$   & $1.95\pm0.35$ \\
Cross  & Fusion     & $8.29\pm0.46$   & $4.43\pm0.29$   & $2.14\pm0.24$   & $1.40\pm0.26$ \\

\bottomrule
\end{tabular}}
\end{table}

It can be observed that the largest degradation of the results occurred when using shallow models occurs in the \crosseyed database.
In all cases, the \gls{vis} against \gls{vis} comparison reports the best results and it is the scenario where it presents the lowest degradation of the response in the different depths of the model.

As shown in the \gls{nir} against \gls{nir} and Cross-spectral results in the \crosseyed database, some \gls{eer} values in the fusion of traits is higher than the ones using information from the periocular region only. 
This behavior is due to the weight used in the fusion of features where the low discrimination of the iris region penalizes and degrades the fused matching score, as we discuss in Section~\ref{sec:fusionsize}.

The experiments performed by Nguyen et al.~\cite{Nguyen2018} show that features extracted from intermediate layers of the networks achieved better results compared to deep layer representations.
However, our results report lower \gls{eer} rates using features extracted from deeper layers.
It is important to point out that in~\cite{Nguyen2018} the ResNet152 model (i.e., a deeper model than ResNet50, used in our work) was employed.
The same behavior can be observed in work by Henandez-Diaz et al.~\cite{Diaz2019}, were the authors stated that features extracted from the intermediate layers of the ResNet-101 model reported the best results.
Thus, the deepest layer reported in this work is approximately at the same depth as the intermediate layer reported by Nguyen et al.~\cite{Nguyen2018} and by Hernandez-Diaz et al.~\cite{Diaz2019}.
In another work, Hernandez-Diaz et al.~\cite{Diaz2018} reported that using the ResNet50 model, representations from the intermediate layers achieved better results in the UBIPr Periocular database~\cite{Padole2012}.
Oppositely, in this work, periocular representations extracted from the last layer of the ResNet50 model achieved the best results.
Notice the UBIPr database has some larger images (from $501\times401$ pixels ($8$m) to $1001\times801$ ($4$m)) than \polyu and \crosseyed databases and also the periocular region is more extensive, containing eyebrows information, which can explain why a shallow model can extract more discriminant features from the intermediate layers, in this case.

As described in~\cite{Wang2019}, a disadvantage of the VGG16 model, when compared to ResNet, is its larger number of trainable parameters ($98.6$M, when compared to their CNN with SD methodology $0.6$M).
As before stated, in our case the best responses were observed when using the ResNet50 model, which after the modifications has $24.1$M (four times lower compared to VGG16).
As shown in Table~\ref{tab:depth}, smaller networks in terms of depth lead to increasingly high losses in performance, however also decreasing  nearly 10M training parameters, which can be an interesting solution for embedded systems and other cases where the computational complexity might be a concern. 
The ResNet with $12$ layers has more trainable parameters than the other models, since it considers an input image of $28\times28$ pixels and $128$ filters.
In addition, its convolutional part is connected with a fully connected layer containing $256$ neurons added for reduction of feature dimensionality.

\subsection{Subjective evaluation}
\label{ssec:subjectiveevaluation}

\begin{figure}[!ht]
\centering
\begin{tabular}{cccc}
    \multicolumn{0}{c}{Worst Genuine} \\

   	{\includegraphics[width=0.22\columnwidth]{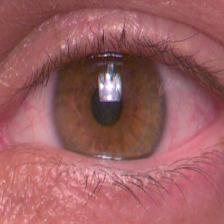}}
    {\includegraphics[width=0.22\columnwidth]{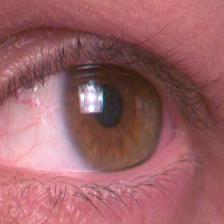}}
    \hspace{1.5mm}
    
	{\includegraphics[width=0.22\columnwidth]{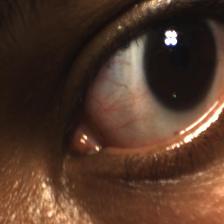}}  {\includegraphics[width=0.22\columnwidth]{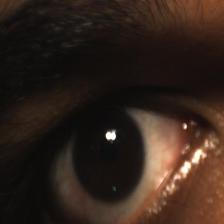}} 
    \vspace{1.5mm} \\
    
    {\includegraphics[width=0.22\columnwidth]{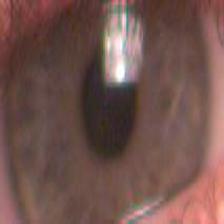}}
    {\includegraphics[width=0.22\columnwidth]{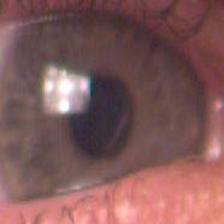}}
    \hspace{1.5mm}

    {\includegraphics[width=0.22\columnwidth]{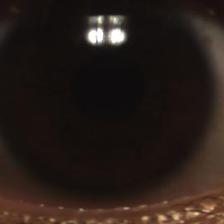}}
    {\includegraphics[width=0.22\columnwidth]{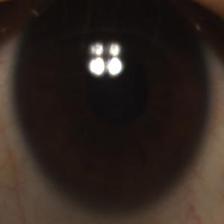}} 
    \vspace{1.5mm} \\

    \multicolumn{0}{c}{Best Impostor} \\

    {\includegraphics[width=0.22\columnwidth]{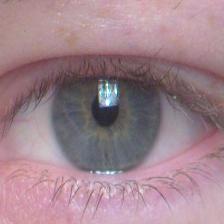}}
    {\includegraphics[width=0.22\columnwidth]{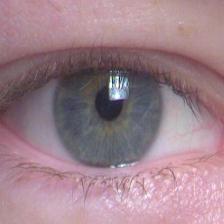}}
    \hspace{1.5mm}

    {\includegraphics[width=0.22\columnwidth]{figs/PolyU/Perioc/C120I12P1.jpg}}
    {\includegraphics[width=0.22\columnwidth]{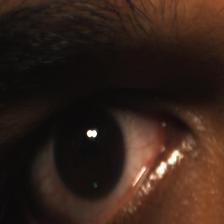}}
	\vspace{1.5mm} \\
    
	{\includegraphics[width=0.22\columnwidth]{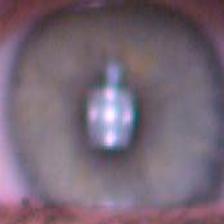}}
    {\includegraphics[width=0.22\columnwidth]{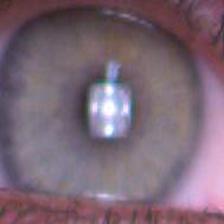}}
    \hspace{1.5mm}
    
    {\includegraphics[width=0.22\columnwidth]{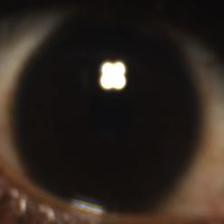}}
    {\includegraphics[width=0.22\columnwidth]{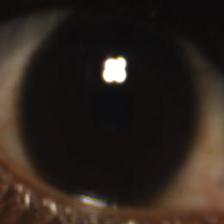}}
    \vspace{1.5mm} \\
 
\end{tabular}
\caption{Pairwise comparison errors in the \gls{vis} against \gls{vis} scenario on \crosseyed (left) and \polyu (right) databases. Periocular and iris matching modalities are presented at Top and Bottom rows, respectively.}
\label{fig:pairwiseerrors}
\end{figure}

In order to provide some insight about the weaknesses of the solutions proposed in this paper, and also to provide a basis for subsequent improvements in the technology, this section highlights some notable cases of image pairwise comparisons that led to the best/worst performance (using the closed-world protocol). Results are shown in Figure~\ref{fig:pairwiseerrors}, grouped into the worst genuine (when the system rejected a true matching) and the best impostors (when the system accepted a false matching)  comparisons.

Although Figure~\ref{fig:pairwiseerrors} only shows \gls{vis} images, we noticed that pose and gaze are factors that can lead to matching errors also in \gls{nir} against \gls{nir} and cross-spectral scenarios.
We observe that there were also confusions in images of the same subject but from different classes (left and right eyes) no matter the spectral scenario.
Thus, we believe that it is possible to improve the recognition system accuracy using information based on the angle of the periocular region images and also performing a preprocessing to determine the left and right eyes (i.e., a soft biometrics process).
Also, based on the pairwise comparison errors, we can state that another factor that may improve system accuracy is the process of centralization/resizing of the periocular image based on the iris region size and location, similar to the method proposed by Hernande-Diaz et al. ~\cite{Diaz2018}.

%% file: text/conclusion.tex
\section{Conclusion}
\label{sec:conclusion}

In this work we performed extensive experiments on two databases for both cross-spectral and intra-spectral ocular recognition.
A strategy using methodologies from the literature was applied to reach new state-of-the-art results on both databases.
It shows that there is still room for improvement by applying and merging known methodologies in the literature to surpass cross-spectral ocular recognition.

We also discuss how deep representations from the iris and ocular region (extracted using VGG16 and ResNet50 architectures) can be fused to improve the recognition performance on the ambitious cross-spectral recognition problem.
We used \gls{cnn} models that were pre-trained for face recognition, and fine-tuned each one for a specific biometric modality: iris and periocular.
A single model for each trait was trained for the feature extraction using \gls{nir} and \gls{vis} images.
The matching phase, on a verification mode, was performed using the cosine metric.
In order to provide a fair comparison with the state-of-the-art approaches, we used the closed-world protocol.
However, we also reported results on the open-world protocol to evaluate the robustness of the proposed methodology.

Our experiments showed that the models learned on the ResNet-50 architecture reported best results in terms of \gls{eer} than its VGG counterpart, both in the \polyu and \crosseyed databases. Interestingly, we note that even this simple processing chain was observed to advance the state-of-the-art results in both datasets.

Overall, in most of the experiments, features taken from the periocular region were observed to provide better performance than iris features, with the fusion of these two modalities improving the \gls{eer} value and decidability index than the best individual trait. 

In a complementary way, we analyzed the impact of the feature vector size and the Iris and Periocular weights used for trait representation fusion, and how the recognition performance varies with respect to the depth of the models used for feature extraction, i.e., by using intermediate layers of the ResNet50 model to take the feature sets used in the matching phase.

Finally, subjective analysis of the best/worst false genuine and true impostors image pairwise comparisons was also performed, showing that factors such as angles of image capture may interfere in the accuracy of the recognition system.
In this direction, we plan for future works to investigate how to build representation taking into account eye gaze and pose.